\documentclass[a4paper]{article}

\usepackage{INTERSPEECH2021}

\usepackage[export]{adjustbox}
\usepackage{url}
\usepackage{amsmath}
\usepackage{multirow}

\title{On Sampling-Based Training Criteria for Neural Language Modeling}
\name{Yingbo Gao$^{1,2}$, David Thulke$^{1,2}$, Alexander Gerstenberger$^{1}$, Khoa Viet Tran$^{1}$,\\ Ralf Schlüter$^{1,2}$, Hermann Ney$^{1,2}$}
\address{$^1$Human Language Technology and Pattern Recognition Group, Computer Science Department\\RWTH Aachen University, 52074 Aachen, Germany $^2$AppTek GmbH, 52062 Aachen, Germany}
\email{\{ygao|thulke|schlueter|ney\}@cs.rwth-aachen.de, \{alexander.gerstenberger|khoa.tran\}@rwth-aachen.de}

\begin{document}

\maketitle

\begin{abstract}
As the vocabulary size of modern word-based language models becomes ever larger, many sampling-based training criteria are proposed and investigated.
The essence of these sampling methods is that the softmax-related traversal over the entire vocabulary can be simplified, giving speedups compared to the baseline.
A problem we notice about the current landscape of such sampling methods is the lack of a systematic comparison and some myths about preferring one over another.
In this work, we consider Monte Carlo sampling, importance sampling, a novel method we call compensated partial summation, and noise contrastive estimation.
Linking back to the three traditional criteria, namely mean squared error, binary cross-entropy, and cross-entropy, we derive the theoretical solutions to the training problems.
Contrary to some common belief, we show that all these sampling methods can perform equally well, as long as we correct for the intended class posterior probabilities.
Experimental results in language modeling and automatic speech recognition on Switchboard and LibriSpeech support our claim, with all sampling-based methods showing similar perplexities and word error rates while giving the expected speedups.
\end{abstract}

\section{Introduction}

Enjoying the benefit of large amounts of text-only training data, language models (LMs) remain an important part of the modern automatic speech recognition (ASR) pipeline \cite{liu2014efficient, irie2016lstm, beck2019lstm}.
However, the large quantity of available data is a double-edged sword, posing real challenges in training.
For example, the popular BERT model \cite{devlin-etal-2019-bert} and the recent GPT-2 and GPT-3 models \cite{radford2019language, brown2020language} have millions of parameters and are trained on billions of tokens.
For BERT and GPT, both systems use byte pair encoding \cite{sennrich2015neural} to mitigate the problem of potentially very large vocabularies.
However, for ASR, it is not uncommon for LMs to operate on the word level with a vocabulary size in the order of several hundred thousands \cite{wang2020transformer, luscher2019rwth}.

In order to train the neural LMs more efficiently, many speedup methods are proposed.
To name a few: hierarchical softmax is used in \cite{mikolov2013efficient}, which changes the flat classification over all words to a series of binary decisions to arrive at the correct word; negative sampling (NS) method in \cite{NIPS2013_9aa42b31} sums over a few sampled words instead of the full vocabulary; the noise contrastive estimation (NCE) method is proposed in \cite{gutmann2010noise} and adapted later in \cite{mnih2012fast} for efficient estimation of LMs.
Note that this is in no way an exhaustive enumeration of the ideas and methods, because there exist other works that introduce interesting concepts to address the problem, e.g. the Monte Carlo (MC) sampling and importance sampling (IS) discussed in \cite{bengio2003quick}.
For an overview of approximations to softmax, we refer the readers to a comprehensive blog post by Sebastian Ruder \cite{ruder2016wordembeddingspart2}.

Among these methods, we find an interesting appreciation for NCE.
In \cite{goldberger2018self}, the authors discuss the self normalization properties of models trained with NCE.
In another line of work that maximizes mutual information for representation learning \cite{oord2018representation, poole2019variational, tschannen2019mutual, kong2019mutual}, the NCE concept is frequently used.
In a preprint note by Chris Dyer \cite{dyer2014notes}, a short conclusion is drawn ``Thus, if your goal is language modeling, you should use NCE; if your goal is word representation learning,
you should consider both NCE and negative sampling."
As a result, if one overlooks the math and takes it for granted, methods like NS may not sound very attractive for language modeling.

In this paper, we start from three fundamental criteria, namely mean squared error (MSE), binary cross-entropy (BCE), and cross-entropy (CE).
We explicitly write out the sampling-based versions of them.
Then, we derive the theoretical optimums of the model, and show that although for some sampling-based criteria we may not directly obtain the original class posterior probabilities from the model outputs, because there is a one-to-one mapping between the optimum and the posterior, we can correct the model outputs and obtain the desired probabilities.
By doing so, the model also has self-normalization properties and gives reasonable performances.
Note that we are not here to argue that any of the cited work is fundamentally wrong, our goal is simply to raise awareness that models trained with sampling-based criteria other than NCE can also perform well, given enough care.

Our contribution can be summarized into two points:
\begin{itemize}
    \item First, we examine various sampling methods systematically, linking back to traditional criteria MSE, BCE and CE, and derive optimums for them.
    \item Second, we show from both a theoretical and a practical perspective, that all these sampling methods under consideration work, giving similar perplexities (PPLs), word error rates (WERs) and speedups.
\end{itemize}

\section{Related Work}

Neural LMs are commonly used in second-pass rescoring \cite{liu2014efficient, irie2016lstm, kumar2017lattice, li2021parallelizable} or first-pass decoding \cite{beck2019lstm} in ASR systems.
While for conventional research-oriented datasets like Switchboard the word-level vocabulary size is several dozens of thousands, for larger systems, especially commercially available systems, the vocabulary size can often go up to several hundred thousand.
Numerous methods to speed up the training (and potentially testing) of word-level LMs are proposed in the past decades \cite{mikolov2013efficient, NIPS2013_9aa42b31, gutmann2010noise, mnih2012fast, bengio2003quick, ruder2016wordembeddingspart2}.
These methods either exploit the statistical structure or perform sampling in the large target vocabulary.
Among the sampling methods, NCE enjoys special appreciation \cite{oord2018representation, poole2019variational, tschannen2019mutual, kong2019mutual, dyer2014notes}, whose self-normalization property is also examined \cite{goldberger2018self}.
For self-normalization and variance regularization, it is shown that explicit additional losses can also be added to the cross-entropy training criterion \cite{chen2016efficient}.
Traditionally, MSE, BCE, and CE are three training criteria that give the correct class posterior probabilities \cite{golik2013cross, golik2020:phd}.
In this paper, we make a connection between sampling-based criteria and these traditional ones and show that as long as one corrects for the intended class posterior probabilities, these sampling methods all show similar PPLs and WERs, while giving the expected speedups.

\section{Sampling-Based Training Criteria}

In this section, we formally define the training criteria.
To clarify the notations, in the following sections, $n$ is a running index in the number of word positions $N$, $c$, $c'$ and $\Tilde{c}$ are running indices in the target vocabulary $C$, which is supposedly very large.
The context or history for next word prediction is denoted with $x$, and the Kronecker delta $\delta$ is used to decide the identity of model prediction and ground truth.
We use $\theta$ for model parameters, $q$ for model outputs, $p$ for class posterior probabilities, and $\hat{q}$ to represent the derived optimum.
When model outputs are explicitly normalized, $q(c|x)$ is used, otherwise, $q(x,c)$ is used.
$F$ represents the training criterion to maximize.
Lastly, the distribution from which we sample is denoted with $\mathcal{D}$, and $k$ is a running index in the number of samples $K$.
Due to the limited pages, we only show the criterion definition and the optimum, and comment on the important steps in derivations in this paper.

\subsection{Traditional Criteria}

\subsubsection{Mean Squared Error (MSE)}

MSE is a classic training criterion commonly used for regression problems.
Intuitively, it corresponds to ``counting the errors", but in the continuous sense.
\begin{align}
F_\text{MSE}(\theta) &:= - \frac{1}{N} \sum_{n}^{N} \sum_{c}^{C} \left( q_{\theta} (x_{n}, c) - \delta(c_{n}, c) \right)^2\\
&\hspace{-3em}\implies \hat{q}_{\theta} (x,c) = p(c|x)
\end{align}
To obtain the optimum, one could rewrite the summation in $N$ into a summation in $x$, expand the squared term, and single out the terms related to $q_\theta$.
By definition, $q$ can be unbounded, but it is common to parametrize $q$ to be positive.
In our preliminary experiments, further constraining $q$ to be between zero and one with sigmoid slightly boosts the performance.
However, although we spend a great amount of effort to tune the MSE-based models\footnote{For example, initializing the model with parameters from a converged NCE-trained model, changing optimizers and grid-searching over the learning rates. We also find that down-scaling the penalization MSE has on the rival classes improves the PPL a little, but similar experiments with BCE or CE do not give any improvements. Although authors of \cite{hui2020evaluation} claim that MSE should be preferred, we argue that for large number of target classes, it is hard for MSE to converge to the optimum with our current training recipes. Our empirical experience also agrees with that in \cite{golik2013cross}.}, the PPLs are still much worse than BCE and CE, therefore we only describe the MSE criterion here for the sake of completeness and not mention it later.

\subsubsection{Binary Cross Entropy (BCE)}

BCE is another traditional training criterion.
The motivation of BCE can be summarized as to ``encourage the correct predictions and discourage the wrong predictions".
\begin{align}
F_\text{BCE}(\theta) &:=\\\nonumber
&\hspace{-4em}\frac{1}{N} \sum_{n=1}^{N} \Big( \log q_{\theta} (x_{n}, c_{n}) + \sum_{c' \neq c_n}^{C} \log (1 - q_{\theta} (x_{n}, c')) \Big)\\
&\hspace{-3em}\implies \hat{q}_{\theta} (x, c) = p(c|x)
\end{align}
Here, divergence inequality can be used for derivation.
Note that $q$ is required to be bounded in (0, 1) and it is commonly done via a sigmoid operation.

\subsubsection{Cross Entropy (CE)}

CE is arguably the most commonly used criterion nowadays and finds its roots in information theory and probabilistic theory.
Intuitively, CE ``encourages the model probability on the target word to be more exactly correct".
\begin{align}
F_\text{CE}(\theta) &:= \frac{1}{N} \sum_{n=1}^{N} \log \frac{\exp q_\theta(x_n, c_n)}{\sum_{c'=1}^C \exp q_\theta(x_n, c')}\\
&\hspace{-3em}\implies \frac{\exp \hat{q}_\theta(x, c)}{\sum_{c'=1}^C \exp \hat{q}_\theta(x, c')} = p(c|x)
\end{align}
Again, applying divergence inequality here would give us the optimum.
We explicitly write out the softmax operation in this case to highlight the summation in the big vocabulary $C$ in the denominator.
In this case, $q$ is unbounded and the softmax guarantees the normalized property.

\subsection{Sampling-Based Criteria}

According to the previous discussion about MSE, BCE, and CE, we see a summation in $C$ in all three cases.
This summation can be viewed as an expectation of some quantity $Q_c$ under the uniform distribution $\frac{1}{C}$: $\sum_c^C Q_c = C \sum_c^C \frac{1}{C} Q_c = C \mathbb{E}(Q_c)$.
Approximating this expectation is the core concept of the sampling methods.

\subsubsection{Monte Carlo Sampling (MCS)}

MCS approximates an expectation by some sample mean.
Specifically, instead of summing over $C$, we sum over $K$ random samples where $\Tilde{c}_k \sim \mathcal{D}$.
NS in \cite{mikolov2013efficient} is a prominent example of MCS.
\begin{align}
F_\text{BCE-MCS}(\theta) &:=\\\nonumber
&\hspace{-4em}\frac{1}{N} \sum_{n=1}^{N} \Big( \log q_{\theta} (x_{n}, c_{n}) + \sum_{k=1}^{K} \log (1 - q_{\theta} (x_{n}, \Tilde{c}_k)) \Big) \\
&\hspace{-3em}\implies \hat{q}_{\theta} (x, c) \approx (1 + \frac{K \mathcal{D}(c)}{p(c|x)})^{-1}\\
F_\text{CE-MCS}(\theta) &:=\\\nonumber
&\hspace{-4em}\frac{1}{N} \sum_{n=1}^{N} \Big( q_\theta(x_n, c_n) - \log \sum_{k=1}^K \exp q_\theta(x_n, \Tilde{c}_k) \Big)\\
&\hspace{-3em}\implies \frac{\mathcal{D}(c) \exp \hat q_\theta(x, c)}{\sum_{c'=1}^C \mathcal{D}(c') \exp \hat q_\theta(x, c')} \approx p(c|x)
\end{align}

\subsubsection{Importance Sampling (IS)}

IS rewrites the expectation by introducing another distribution other than the uniform distribution.
In our case, this new distribution is the noise distribution $\mathcal{D}$.
\begin{align}
F_\text{BCE-IS}(\theta) &:=\\\nonumber
&\hspace{-4em}\frac{1}{N} \sum_{n=1}^{N} \Bigg( \log q_{\theta} (x_{n}, c_{n}) + \sum_{k=1}^{K} \frac{\log (1 - q_{\theta} (x_{n}, \Tilde{c}_k))}{K\mathcal{D}(\Tilde{c}_k)}  \Bigg) \\
&\hspace{-3em}\implies \hat{q}_{\theta} (x, c) \approx (1 + \frac{1}{p(c|x)})^{-1}\\
F_\text{CE-IS}(\theta) &:=\\\nonumber
&\hspace{-4em}\frac{1}{N} \sum_{n=1}^{N} \Bigg( q_\theta(x_n, c_n) -\log \sum_{k=1}^K \frac{\exp q_\theta(x_n, \Tilde{c}_k)}{K\mathcal{D}(\Tilde{c}_k)} \Bigg) \\
&\hspace{-3em}\implies \frac{ \exp \hat q_\theta(x, c)}{\sum_{c'=1}^C  \exp \hat q_\theta(x, c')} \approx p(c|x)
\end{align}

\subsubsection{Compensated Partial Summation (CPS)}

CPS comes from a very simple motivation.
If we replace the summation of $C$ terms with a sub-summation of $K$ terms, maybe we should compensate the partial summation by $\frac{C}{K}.$
Essentially, this is still a Monte Carlo method with some correction term $\alpha = \frac{C}{K}$.
\begin{align}
F_\text{BCE-CPS}(\theta) &:=\\\nonumber
&\hspace{-4em}\frac{1}{N} \sum_{n=1}^{N} \Big( \log q_{\theta}(x_{n}, c_{n}) + \alpha \sum_{k=1}^{K} \log (1 - q_{\theta}(x_{n}, \Tilde{c}_k)) \Big) \\
&\hspace{-3em}\implies \hat{q}_{\theta} (x, c) \approx (1 + \frac{\alpha K \mathcal{D}(c)}{p(c|x)})^{-1}\\
F_\text{CE-CPS}(\theta) &:=\\\nonumber
&\hspace{-4em}\frac{1}{N} \sum_{n=1}^{N} \Big( q_\theta(x_n, c_n) - \log \alpha \sum_{k=1}^K \exp q_\theta(x_n, \Tilde{c}_k) \Big) \\
&\hspace{-3em}\implies \frac{\mathcal{D}(c) \exp \hat q_\theta(x, c)}{\sum_{c'=1}^C \mathcal{D}(c') \exp \hat q_\theta(x, c')} \approx p(c|x)
\end{align}

\subsubsection{Noise Contrastive Estimation (NCE)}

NCE changes the task of next-word prediction to a binary classification task of telling true samples from noisy ones.
Originally, NCE is proposed in the context of BCE.
Due to the property that the class posterior probabilities can be obtained directly from the model output, NCE is one of the most popular methods seen in literature \cite{goldberger2018self, oord2018representation, poole2019variational, tschannen2019mutual, kong2019mutual, dyer2014notes, chen2016efficient}.
\begin{align}
F_\text{BCE-NCE}(\theta) &:= \frac{1}{N} \sum_{n=1}^{N} \Bigg( \log \frac{q_{\theta} (x_{n}, c_{n})}{q_{\theta} (x_{n}, c_{n}) + K \mathcal{D}(c_n)} \\\nonumber
&\hspace{-4em} + \quad \sum_{k=1}^{K} \log (1 - \frac{q_{\theta} (x_{n}, \Tilde{c}_k)}{q_{\theta} (x_{n}, \Tilde{c}_k) + K \mathcal{D}(\Tilde{c}_k)}) \Bigg)\\
&\hspace{-3em}\implies \hat{q}_{\theta} (x, c) \approx p(c|x)\\
F_\text{CE-NCE}(\theta) &:= \frac{1}{N} \sum_{n=1}^N \Bigg( \frac{q_\theta(x_n, c_n)}{q_\theta(x_n, c_n) + K \mathcal{D}(c_n)} \\\nonumber
&\hspace{-4em} - \quad \log \sum_{k=1}^K \exp (\frac{q_\theta(x_n, \Tilde{c}_k)}{q_\theta(x_n, \Tilde{c}_k) + K \mathcal{D}(\Tilde{c}_k)}) \Bigg) \\
&\hspace{-3em}\implies \frac{\mathcal{D}(c) \exp ( \frac{q_\theta(x,c)}{q_\theta(x,c) + K \mathcal{D}(c)})}{\sum_{c'} \mathcal{D}(c') \exp ( \frac{q_\theta(x,c')}{q_\theta(x,c') + K \mathcal{D}(c')})} \approx p(c|x)
\end{align}

\subsubsection{Notes on Derivation and Implications}

The term $K \mathcal{D}(c)$ often shows up in our derivation due to rewriting the summation in $K$ into a summation in $C$: $\sum_n^N\sum_k^K Q_{n, k} \approx \sum_n^N\sum_c^C K \mathcal{D}(c) Q_{n, c}$.
Approximately, the number of terms that show up in the two summations should be equal, when given $n$ and $K$ is large enough.
This trick is used in many of the derivations.
Note that the model outputs $q$ have different constraints in different criteria, the logits we plug into the criteria are activated according, e.g. applying sigmoid if $q$ is bounded between zero and one.

From the derived optimums, if the model output $q(x, c)$ is not strictly the intended class posterior probability $p(c|x)$ (unlike in the case of three traditional criteria and BCE-NCE), we confirm the statement in \cite{dyer2014notes} that such models are not directly applicable for language modeling.
However, for all cases, it is clear that there is a one-to-one mapping between $q$ and $p$.
Therefore, given each model output $q(x, c)$, we can calculate the desired $p(c|x)$ (optionally querying the noise distribution $\mathcal{D}$), and use this quantity for rescoring in ASR.

\section{Experiments}

\subsection{Experimental Setup}

For the experimental validation, we train word-based LMs on LibriSpeech and SwitchBoard and evaluate the resulting models in well-tuned ASR systems with second-pass lattice rescoring. The vocabulary contains about 200k words for LibriSpeech and about 30k words for SwitchBoard, respectively.

For all training criteria, we make use of the same model architecture\footnote{Ideally, one should tune the hyperparameters for each training criterion individually. Given the computational and time constraints, we believe that this is still a reasonable approach.}.
For LibriSpeech, our LMs make use of the Transformer \cite{NIPS2017_3f5ee243} architecture, which is motivated by recent state-of-the-art results outperforming LSTM-based LMs on this task by a large margin \cite{luscher2019rwth,irie19:trafolm}.
We use 42 layers, 512 input embedding dimension, 2048 feed-forward dimension, 8 attention heads, and 512 residual and key/query/value dimension.
We do not use a positional encoding, following the architecture presented in \cite{irie19:trafolm}.
For SwitchBoard, we use LSTM LMs with 2 layers and 2048 hidden units, and 128 embedding dimension.

For sampling-based approaches, we sample 8192 samples from log-uniform noise distributions, which performs well in our preliminary experiments.
The noise samples are shared over the whole batch for computational efficiency in all cases \cite{zoph2016simple, jozefowicz2016exploring}.
The parameters are learned using stochastic gradient descent with a learning rate of 1.
The global gradient norms are clipped to 1.
We implement our models using the open-source toolkit \mbox{RETURNN}\footnote{Example config is available at \url{https://git.io/Jnq3Q}.} \cite{zeyer2018:returnn}, based on \mbox{TensorFlow} \cite{tfshort}.

The acoustic models are based on hybrid hidden Markov models.
For detailed information on training setups, including discriminative training approaches, adaptation, and model architectures we refer the reader to \cite{kitza19:interspeech} for the SwitchBoard and \cite{luscher2019rwth} for the LibriSpeech systems.
For LibriSpeech the lattices are generated using a well-tuned LSTM LM in the first pass \cite{beck2019lstm}.
For SwitchBoard, the lattices are generated using a 4-gram Kneser-Ney LM \cite{kitza19:interspeech}.
We interpolate the LSTM LMs with count-based LMs for evaluation.
We report on clean and other test sets for LibriSpeech and on SwitchBoard (SWB) and CallHome (CH) Hub5'00 test sets for SwitchBoard.
During LM training we use a separate validation dataset.
PPLs and WERs are always obtained with proper normalization over the full vocabulary unless noted otherwise.
The reported average training time per batch is obtained on NVIDIA GeForce GTX 1080 Ti GPUs, with a batch size of 64 for LibriSpeech and 32 for Switchboard respectively.

\subsection{Main Results}

Below we present the main results.
Table \ref{tab:baselines} gives the PPLs and WERs of our baseline models \cite{beck2019lstm, kitza19:interspeech}.
In our opinion, these are competitive systems and serve as reasonable baselines for our purpose of judging different sampling methods.

\begin{table}[!h]
\setlength{\tabcolsep}{0.3em}  
\centering
\caption{\it Baseline PPLs and WERs.}
\label{tab:baselines}
\begin{tabular}{|c|c|c|c|c|c|}
\hline
dataset & model & PPL & \multicolumn{3}{c|}{WER} \\ \hline\hline
\multirow{2}{*}{LibriSpeech} & \multirow{2}{*}{LSTM} & \multirow{2}{*}{64.3} & - & clean & other \\ \cline{4-6} 
 &  &  & - & 2.6 & 5.8 \\ \hline\hline
\multirow{2}{*}{SwitchBoard} & \multirow{2}{*}{4-gram} & \multirow{2}{*}{74.6} & ALL & SWB & CH \\ \cline{4-6} 
 &  &  & 11.8 & 8.1 & 15.4 \\ \hline
\end{tabular}
\vspace{-0.5em}
\end{table}

In Table \ref{tab:lrs}, we present the PPLs and WERs of the sampling-based models on the LibriSpeech dataset.
Since we sample 8k samples out of a 200k vocabulary, for all sampling methods under consideration, there is a significant relative training time speedup of over 40\%, which is expected.
Because of limited computational resources and our purpose not being to obtain the best trade-off between speed and quality, we do not sweep over different sampling sizes.
Note that we do not include CE-NCE results due to some numerical problems.
In terms of PPLs and WERs, we see a consistent improvement over the baseline, and we attribute this to using the Transformer and not the LSTM architecture \cite{luscher2019rwth,irie19:trafolm}.
Comparing different sampling methods, we see a small variation in PPLs and even less in WERs.
This justifies our statement that all these sampling methods work equally well when the model outputs are corrected accordingly.

\begin{table}[!ht]
\setlength{\tabcolsep}{0.3em}  
\centering
\caption{\it Sampling-Based Transformer LMs on LibriSpeech.}\label{tab:lrs}
\begin{tabular}{|c|c|c|c|c|c|}
\hline
\multirow{2}{*}{criterion} & \multirow{2}{*}{sampling} & \multirow{2}{*}{\begin{tabular}[c]{@{}c@{}}train time\\ (ms/batch)\end{tabular}} & \multirow{2}{*}{ PPL } & \multicolumn{2}{c|}{WER} \\ \cline{5-6} 
 &  &  &  & clean & other \\ \hline\hline
\multirow{5}{*}{BCE} & - & 0.358 & 58.5 & 2.5 & 5.4 \\ \cline{2-6} 
 & MCS & 0.213 & 58.0 & 2.6 & 5.4 \\ \cline{2-6} 
 & IS & 0.206 & 58.4 & 2.6 & 5.5 \\ \cline{2-6} 
 & CPS & 0.205 & 58.4 & 2.5 & 5.4 \\ \cline{2-6} 
 & NCE & 0.206 & 57.9 & 2.5 & 5.4 \\ \hline\hline
\multirow{5}{*}{CE} & - & 0.302 & 57.7 & 2.5 & 5.4 \\ \cline{2-6} 
 & MCS & 0.206 & 57.9 & 2.5 & 5.4 \\ \cline{2-6} 
 & IS & 0.201 & 58.7 & 2.5 & 5.4 \\ \cline{2-6} 
 & CPS & 0.203 & 62.2 & 2.5 & 5.4 \\ \hline
\end{tabular}
\vspace{-0.5em}
\end{table}

In Table \ref{tab:swb}, similar results are presented for the SwitchBoard dataset.
In this case, all sampling-based methods give consistent relative training time speedup of over 20\%.
Considering that here 8k samples are drawn out of a 30k vocabulary, instead of 8k being drawn from a 200k vocabulary like that for LibriSpeech, the relative speedup being smaller is thus expected.
PPL-wise, significant improvements over count-based baseline LM are achieved, which is also confirmed numerous times in different works on different datasets.
The CE baseline is slightly better than the sampling-based LMs because our training recipe is well-tuned for the CE baseline.
Comparing different sampling-based criteria, NCE and IS are slightly better, but the advantage is not large enough to justify one method being strictly superior than another.
In terms of WERs, all sampling methods perform similarly.

\begin{table}[!ht]
\setlength{\tabcolsep}{0.3em}  
\centering
\caption{\it Sampling-Based LSTM LMs on Switchboard.}\label{tab:swb}
\begin{tabular}{|c|c|c|c|c|c|c|}
\hline
\multirow{2}{*}{criterion} & \multirow{2}{*}{sampling} & \multirow{2}{*}{\begin{tabular}[c]{@{}c@{}}train time\\ (ms/batch)\end{tabular}} & \multirow{2}{*}{PPL} & \multicolumn{3}{c|}{WER} \\ \cline{5-7} 
 &  &  &  & All & SWB & CH \\ \hline\hline
\multirow{5}{*}{BCE} & - & 0.107 & 52.3 & 10.3 & 6.9 & 13.7 \\ \cline{2-7} 
 & MCS & 0.077 & 52.6 & 10.3 & 7.0 & 13.6 \\ \cline{2-7} 
 & IS & 0.079 & 51.5 & 10.3 & 7.0 & 13.7 \\ \cline{2-7} 
 & CPS & 0.079 & 52.4 & 10.3 & 7.1 & 13.6 \\ \cline{2-7} 
 & NCE & 0.078 & 51.4 & 10.3 & 7.0 & 13.6 \\ \hline\hline
\multirow{5}{*}{CE} & - & 0.100 & 49.9 & 10.1 & 6.8 & 13.4 \\ \cline{2-7} 
 & MCS & 0.078 & 52.4 & 10.4 & 7.0 & 13.7 \\ \cline{2-7} 
 & IS & 0.076 & 52.3 & 10.2 & 6.9 & 13.6 \\ \cline{2-7} 
 & CPS & 0.077 & 52.3 & 10.3 & 7.0 & 13.6 \\ \hline
\end{tabular}
\vspace{-1.3em}
\end{table}

While PPLs and WERs reported above are properly normalized and give expected training time speedups, what makes these sampling methods especially attractive is the use case without explicit normalization.
To this end, we look at the BCE sampling variants and directly rescore with the model outputs of these LMs, and report the WERs on Switchboard.
As seen in Table \ref{tab:self-norm}, BCE-IS performs on par with BCE-NCE, as well as the CE baseline shown in Table \ref{tab:swb}, which shows its competitiveness and self-normalization properties.
We notice that when plugging in the log-uniform distribution for $\mathcal{D}$, the BCE-MCS and BCE-CPS performance without normalization can get much worse to around 11.8\% WER, which is similar to the count-based model.
We attribute this to having to query the unreliable noise distribution $\mathcal{D}$ during search (which is not the case for BCE-IS and BCE-NCE) and therefore use a smoothed empirical unigram distribution for $\mathcal{D}$ instead.


\begin{table}[!ht]
\setlength{\tabcolsep}{0.3em}  
\centering
\caption{\it WERs without Explicit Normalization on SwitchBoard.}\label{tab:self-norm}
\begin{tabular}{|c|c|c|c|c|}
\hline
WER & BCE-MCS & BCE-IS & BCE-CPS & BCE-NCE \\ \hline\hline
ALL & 11.0  & 10.2  & 11.3 & 10.2  \\ \hline
SWB & 7.3  & 6.9  & 7.5 & 6.9 \\ \hline
CH & 14.7  & 13.6 & 15.1 & 13.6  \\ \hline
\end{tabular}
\vspace{-0.5em}
\end{table}

\vspace{-1em}

\section{Conclusion}

For language modeling with large vocabularies, we consider different sampling-based training criteria.
We start from three traditional criteria and formulate sampling-based versions of them.
We derive optimums and argue that when model outputs are corrected for the intended class posteriors, these methods perform equally well compared to the popular noise contrastive estimation.
Experimental evidence of perplexity and word error rate results on LibriSpeech and SwitchBoard support our claim.
For direct rescoring without explicit normalization, we show the self-normalization properties of such sampling-based models.

\vspace{-0.4em}

\section{Acknowledgements}

This project has received funding from the European Research Council (ERC) under the European Union’s Horizon 2020 research and innovation programme (grant agreement n\textsuperscript{o}~694537, project ``SEQCLAS"). The work
reflects only the authors' views and the European Research Council Executive Agency (ERCEA) is not responsible for any use that may be made of the information it contains.
Simulations were performed with computing resources granted by RWTH Aachen University under project \texttt{rwth0582}.
We thank Christoph Lüscher for providing the LibriSpeech acoustic model and Eugen Beck for the lattices, and Markus Kitza for the SwitchBoard lattices.

\bibliographystyle{IEEEtran}
\bibliography{mybib}

\end{document}